\documentclass{article}
\pdfoutput=1
\usepackage{arxiv}

\usepackage[utf8]{inputenc} % allow utf-8 input
\usepackage[T1]{fontenc}    % use 8-bit T1 fonts
\usepackage{hyperref}       % hyperlinks
\usepackage{url}            % simple URL typesetting
\usepackage{booktabs}       % professional-quality tables
\usepackage{amsfonts}       % blackboard math symbols
\usepackage{nicefrac}       % compact symbols for 1/2, etc.
\usepackage{microtype}      % microtypography
\usepackage{xcolor}         % colors
\usepackage{graphicx}
\usepackage{amsmath}

\title{Searching for internal symbols underlying deep learning}

\author{%
  Jung H. Lee \\
  Pacific Northwest National Laboratory\\
  Seattle, WA \\
  \texttt{jung.lee@pnnl.gov} \\
  \And
  Sujith Vijayan \\
  School of Neuroscience\\
  Virginia Tech\\
  Blacksburg, VA \\
  \texttt{neuron99@vt.edu} \\
}

\begin{document}

\maketitle

\begin{abstract}
Deep learning (DL) enables deep neural networks (DNNs) to automatically learn complex tasks or rules from given examples without instructions or guiding principles. As we do not engineer DNNs' functions, it is extremely difficult to diagnose their decisions, and multiple lines of studies proposed to explain the principles of their operations. Notably, one line of studies suggests that DNNs may learn concepts, the high level features that are recognizable to humans. In this study, we extend this line of studies and hypothesize that DNNs can develop abstract codes that can be used to augment DNNs' decision-making. To address this hypothesis, we combine foundation segmentation models and unsupervised learning to extract internal codes and identify potential use of abstract codes to make DL's decision-making more reliable and safer. 
\end{abstract}

\section{Introduction}
Deep learning (DL) can automatically wire deep neural networks (DNNs) to perform complex tasks, and DNNs' learning capability allows them to outperform humans in some tasks \cite{Lecun2015, deep_review1, deep_review2}. Notably, instead of expert instructions or feature engineering, DL requires only training examples to solve highly complex problems. However, as DNNs' structures are automatically determined without guiding principles, we do not fully understand their decision-making process. That is, DNNs may be useful black box decision-makers, but without safety measures, deploying them in high-stakes domains (e.g., autonomous cars) comes with unpredictable risks \cite{Lipton2016, highstake3}. For the moment, the current generation of DL does not warrant essential safety to be deployed in safety-critical or high-stakes domains. To address this shortcoming, earlier studies sought potential methods to decode and explain DNNs' decisions. There are multiple lines of research in explainable AI (e.g., see Refs. \cite{molnar2022, e23010018, GradCAM, IntegratedGradient, Gradient2}), but in our study we pay special attention to the one probing correlations between hidden layer representations and inputs or between hidden layer representations and outputs (i.e., decisions). Feature visualization \cite{olah2017feature} and network dissection \cite{Bau2017} sought direct links between inputs and hidden layers, while linear probes investigated correlations between outputs and hidden layers \cite{alain2018understanding, TCAV}. 

It is notable that the mapping between hidden layers and visual features in input layers, in principle, establishes links between high level patterns recognizable to humans (often referred to as `concepts') and hidden neurons. The Test with Concept Activation Vectors \cite{TCAV} explicitly extended this idea to detect concepts encoded in neural networks. The concepts (i.e., perceptible patterns) encoded in neural networks could deepen our understanding of DNNs' operating principles and eventually provide explanations (readable to humans) of their decisions. More importantly, it raises the possibility that DNNs could build some abstract codes in hidden layers, some of which could directly be linked to concepts. In this study, we extend this line of studies by seeking abstract codes, which are not necessarily recognizable to humans, in hidden layers functionally linked to DNNs' decisions. To make a clear distinction between human recognizable concepts (inputs' features) and DNNs' own abstract codes, we will refer to DNNs' abstract codes as `symbols' hereafter. 

Our analysis is inspired by two observations in the literature. First, the hidden layers encode distinct contextual information. Earlier layers encode low-level features such as texture and shapes, whereas late layers encode high-level semantic features. Second, the penultimate layer’s outputs are clustered together according to the labels of inputs. As all hidden layers are trained simultaneously via the same algorithm, it seems natural to assume that all hidden layers may share the same operating principles. More specifically, we assume that hidden layers map inputs with the same contextual information into neighboring outputs, as the penultimate layer maps inputs belonging to the same class onto the same clusters. The penultimate layer encodes the labels, and the earlier hidden layers encode different contexts currently unknown to us. If these functional clusters reflecting contextual information exist in the hidden layers, identifying them can conversely reveal how inputs are progressively encoded. That is, we could uncover the symbols using these functional clusters. 

With this possibility in mind, we seek the functional clusters in the hidden layers to obtain the symbols, the abstract codes of DNNs. We note that visual scenes generally consist of numerous distinct objects, and thus, it is necessary to separate them. To this end, we use segmentation models to locate visual objects of interest and analyze the responses of hidden neurons whose receptive fields cover the locations. Our analysis suggests that these symbols can mediate semantic meanings and allow us to 1) monitor the models' decision-making process, 2) detect abnormal operations related to adversarial perturbations and out-of-distribution (OOD) examples and 3) temporarily learn about OOD examples. Based on these results, we propose that hidden layers' responses can be converted to discrete symbols that can help us build more reliable and safer DNNs.

\section{Extracting symbols underlying DNNs' operations}

To test whether DNNs learn to use symbols to make decisions, we analyzed 5 ImageNet models' hidden layers using a subset (Mixed\_13) of ImageNet \cite{imagenet}, previously curated by a python machine learning library \cite{robustness}, and the hidden layer responses of ResNet50 \cite{resnet} trained on the Oxford-IIIT Pet dataset \cite{iit-pet} . Mixed\_13 contains 78 classes out of 1,000 classes of ImageNet, which belong to one of 13 super-classes (Supplementary Table 1). Mixed\_13 consists of two disjointed `training' and `test' sets. Oxford-IIIT Pet dataset contains 37 classes of cat and dog breeds. To extract the symbols, we assumed that the symbols associated with DNN's decision-making would be observed frequently. For instance, if a symbol is associated with a tiger, a picture of a tiger would elicit the symbol frequently. With this assumption in mind, we presented the inputs from Mixed\_13 and searched for repeatedly occurring hidden layer responses by using unsupervised clustering analysis. 

It should be noted that analyzing hidden layer responses can be challenging for two reasons. First, hidden layer responses are extremely noisy high-dimensional data. A massive number of hidden neurons exist in modern DNNs, and their responses are determined by diverse inputs, making dimension reduction essential (see \cite{rathore2021topoact, purvine2022experimental, alain2018understanding} , for instance). Second, visual scenes often contain multiple objects rather than a single one. For instance, a visual scene may include a tiger or tigers walking along the river, playing with other tigers and/or climbing a tree. Thus, to extract codes related to tigers, we need to isolate regions of interest (ROIs) containing tigers and exclude confounding objects (e.g., tree or river). However, ImageNet training examples are labeled with single class names and do not provide sufficient information to identify ROIs. 

To address these challenges, we first used segmentation models to extract ROIs and recorded hidden layer responses spatially corresponding to ROIs. Then, we used ROI-pooling operation, proposed for object detectors \cite{fastrcnn}, to extract smooth 3-by-3 activation vectors per single feature map. Once the activation vectors were extracted, we conducted clustering algorithms to determine the codes repeatedly occurring with Mixed\_13 (a subset of ImageNets). In section 2.1, we discuss how ROIs are extracted from Mixed\_13. In section 2.2, we present the details of extracting hidden layer responses. In section 2.3, we explain our clustering analysis in detail. 

\subsection{ROIs Identification}
\begin{figure}
  \centering
  \includegraphics[width=1\linewidth]{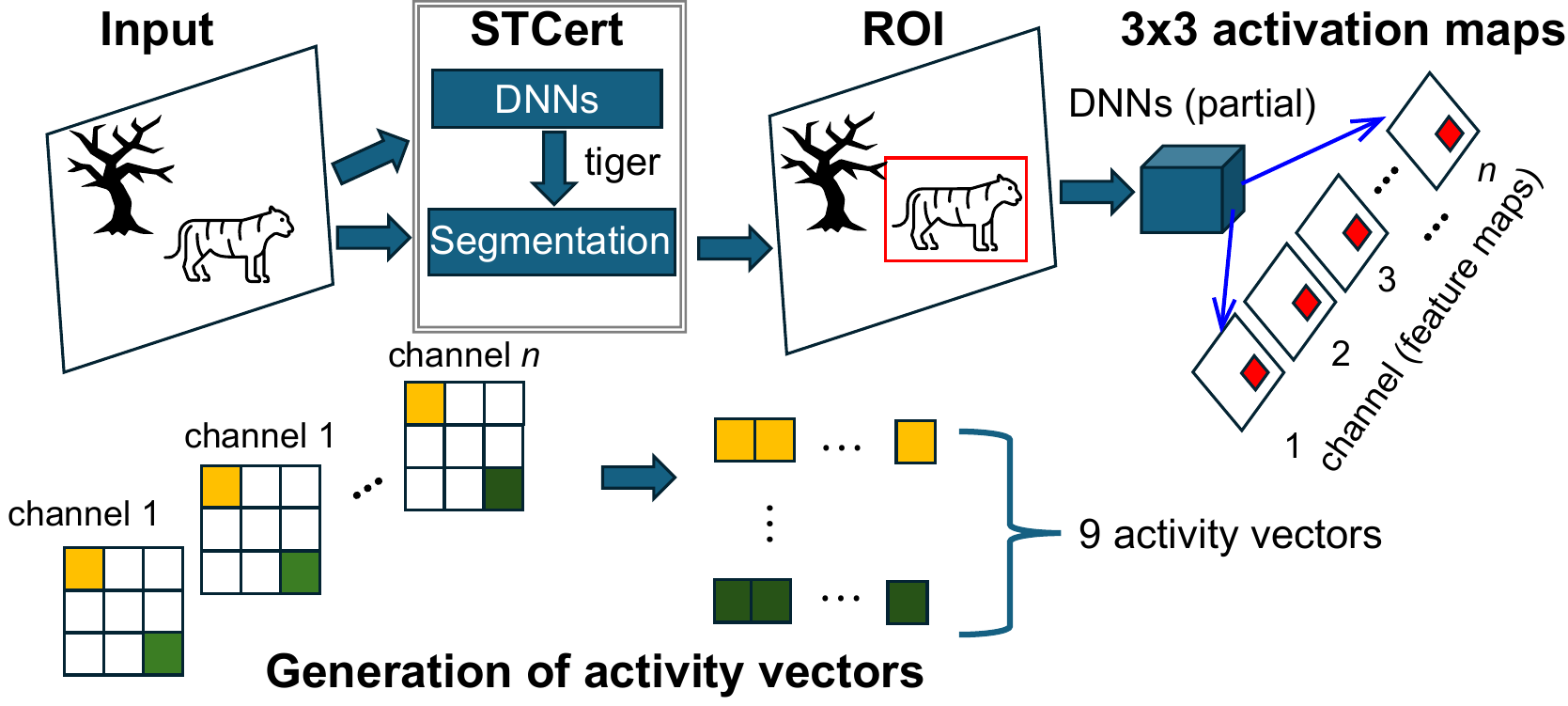}
  \caption{Schematics of STCert \cite{lee2023having}. }
  \label{diagram}
\end{figure}

We note that `Second Thought Certification' (STCert) proposed in an earlier study \cite{lee2023having} can be used to identify ROIs related to visual objects with high accuracy. STCert is a two-step process (Fig. \ref{diagram}). First, it uses foundational segmentation models \cite{liu2023grounding, kirillov2023segany} to detect ROIs related to DNNs' predictions. For instance, if a DNN predicts a tiger, STCert asks the segmentation model to return the bounding box (i.e., ROI) of the tiger in the picture. Second, it crops ROIs and uses them to confirm or reject DNNs' original predictions. STCert provides a more reliable ROI estimates than a single step segmentation, since its segmentation is crosschecked with a classifier. That is, STCert relies on `self-consistency'. As such, we used STCert to identify ROIs in this study to extract ROIs from randomly chosen 200 training examples for each class in Mixed\_13; see Supplementary text  for more detail. Since Mixed\_13 contains 78 classes (Supplementary Table 1), the total number of training examples is 15,600. Among them, we first selected examples, on which DNNs' predictions were correct to minimize the bias induced by the incorrect predictions/decisions of DNNs (i.e., incorrect mapping between ROI and the inputs' labels).  For some of the images, STCert did not find ROIs or found multiple ROIs, which means the actual number of ROIs is determined by a few factors, not just the number of training examples. 

\subsection{Generating hidden layer activity vectors}

Once ROIs were identified, we sampled activations ($H_i$) of hidden neurons, whose receptive fields correspond to ROIs. To reduce noise and filter out non-essential information, we used ROI-pooling  \cite{fastrcnn}  to convert hidden layer activations into 3-by-3 activation maps. That is, the size of hidden layer activation maps can be arbitrary depending on the size of objects, but ROI-pooling always turns them into 3-by-3 activation maps. It should be noted that individual activations in 3-by-3 activation maps reflect pixels at different spatial locations. As we assumed that symbols are collectively encoded neurons that share the same receptive field, we built vector codes (referred to as activity vectors below) by collecting activations at the same location in the maps across channels. For instance, there are 512 channels in ResNet18's layer 4, and we have 9 location-specific activation maps in each channel. Here, we aggregated activations from the same location in the activation map from all 512 channels and then created 9 activity vectors, each of which contains 512 components from the layer per image. 

In this study, we analyzed 4 hidden layers of popular DNNs, ResNet18, ResNet50, VGG19, DenseNet121, and Vision Transformers (ViT) \cite{resnet, vgg19, huang2018densely, vit}. ResNet and DenseNet consist of 4 composite blocks, and their outputs were analyzed. VGG19 contains 5 max pooling layers, whose outputs roughly correspond to ResNet's composite block's outputs. We selected the last 4 max-pooling layers and analyzed them, as they match the layers of ResNet and DenseNet. For ViT consisting of 12 layers, we chose the 4 layers (layer ids: 3, 6, 9, 12) . We used Pytorch \cite{Paszke2017} and Timm \cite{rw2019timm} python deep learning libraries to implement these models. A workstation with Intel Core9 CPU and RTX 4090 was used for all our experiments and analyses.

\subsection{Clustering analysis of hidden layer activity vectors}

We turned to unsupervised clustering analysis to identify representative activity vectors equivalent to symbols in this study. We also note that many activity vectors contain low values. With these low-value activity vectors, clustering algorithms consider the zero-valued vector as the most prominent cluster center, which is not informative. Thus, we removed the vectors consisting of values lower than the mean activity value estimated on each layer. Next, we used UMAP analysis \cite{umap_paper, SMG2020} to further reduce them to 3-dimensional vectors; see Table \ref{tab_cumap} for the parameters used in the analysis. These three dimensional vectors were analyzed to detect symbols via clustering. It should be noted that the exact number of clusters existing in the activity vector space are unknown, and the result of clustering analysis strongly depends on the number of predefined clusters. To address this issue, we used X-means clustering \cite{xmeans} to automatically discover an optimal number of clusters (i.e., the number of symbols). We note that X-means is an extension of Kmeans clustering, which uses Bayesian Information Criterion (BIC) to determine the optimal number of clusters, and a python package `pyclustering' \cite{Novikov2019} was used to implement X-means in this study. In our analysis, we restrict the maximum number of classes to 1,000. Table \ref{tab_symbol} shows the identified number of clusters in 4 layers of DNNs tested. As shown in the table, the optimal numbers of clusters in the early layers are smaller than 1,000, but those in the late layers hit the maximum value, suggesting that homogeneity of activity vectors decreases, as the layers go deeper.

\begin{table}[]
\caption{List of Parameters used in UMAP analysis.}
\label{tab_cumap}
\begin{center}
\begin{tabular}{ll}
\hline
Parameter                         & Value     \\ \hline
n\_neighbors                      & 50        \\
min\_dist                         & 0.1       \\
n\_component & 3         \\
metric                            & Euclidean \\ \hline
\end{tabular}

\end{center}

\end{table}

\section{Symbols Analysis} 

\subsection{Links between symbols and semantic meanings of inputs}
\begin{table}[]
\caption{Number of symbols identified from 4 models. For VGG19, we list 4 last max-pooling layers, and for all other layers, we list 4 outputs of 4 functional blocks.}
\label{tab_symbol}
\begin{center}
\begin{tabular}{ccccc}
\hline
Layer & ResNet18 & ResNet50 & VGG19 & DenseNet121 \\ \hline
1     & 960      & 967      & 837   & 895         \\ \hline
2     & 940      & 992      & 1000  & 1000        \\ \hline
3     & 1000     & 1000     & 1000  & 1000        \\ \hline
4     & 1000     & 1000     & 1000  & 1000        \\ \hline
\end{tabular}
\end{center}
\end{table}

Then, we asked if the identified symbols (i.e., cluster centers) can be linked to semantic meanings of inputs (i.e., labels) by correlating them with inputs' labels. Specifically, for each image ($img_k$), we obtained ROIs ($ROI_m$) and 9 symbols ($S_{mn}$, where $n=1,...,9$). These 9 symbols $S_{mn}$ were correlated with the labels ($l_m$) of $ROI_m$, which were determined by the label of $img_k$ containing $ROI_m$. That is, Correlation Map ($CM(i, j)$) evolves over all identified ROIs obtained from the training examples of Mixed\_13 according to Eq. \ref{eq_cmap}. 
\begin{equation}\label{eq_cmap}
CM(i,j)=
 \begin{cases}
      CM(i, j)+1 & \text{if $i=S_{mn}$, $j=l_m$}\\
      CM(i, j) & \text{otherwise}

    \end{cases}
\end{equation}
, where $n=1,..., 9$, and  $CM(i,j)$ is the component of Correlation map at $i^{th}$ row and $j^{th}$ column; $S_{mn}$ denotes the $n^{th}$ (out of 9) symbol related to $m^{th}$ ROI; where $l_m$ is the label of $m^{th}$ ROI (i.e., $img_k$).

\begin{figure}
  \centering
  \includegraphics[width=1\linewidth]{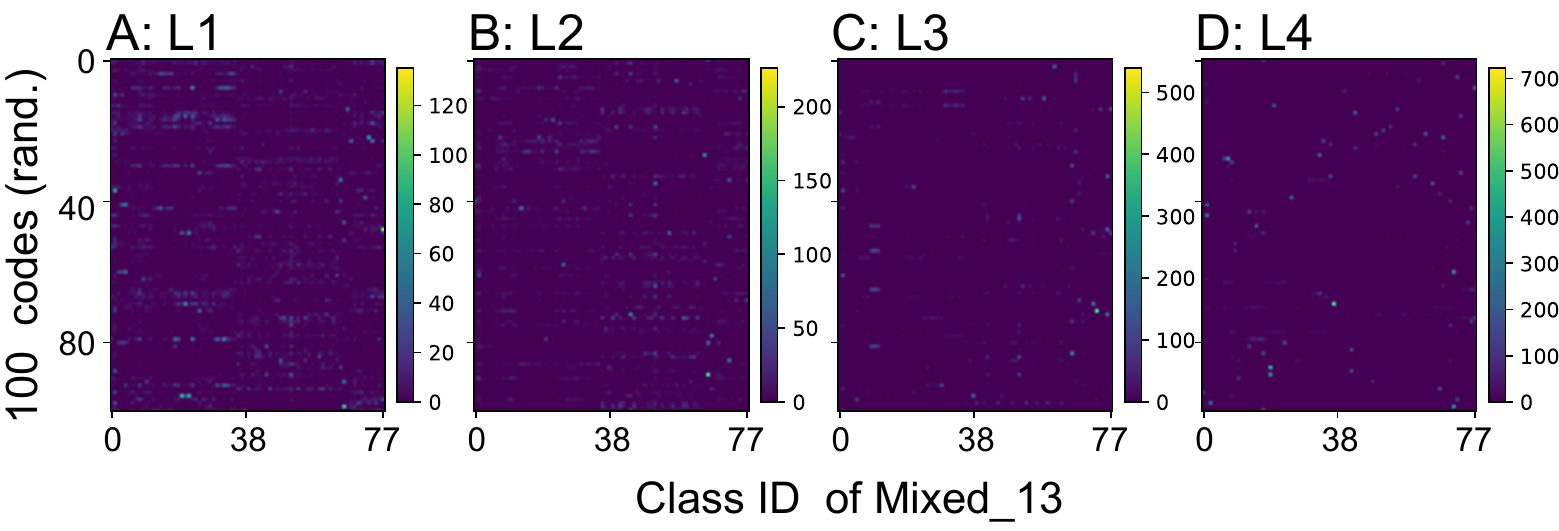}
  \caption{The correlations between the symbols of ResNet18 and classes. We show the correlation maps $CM(i,j)$ between 100 symbols randomly chosen and all 78 classes of Mixed\_13. $y$-axis denotes the indices of symbols, $x$-axis denotes the class. (A)-(D), the correlations observed in layers 1-4, respectively.}
  \label{correlations}
\end{figure}

\begin{figure}
  \centering
  \includegraphics[width=1\linewidth]{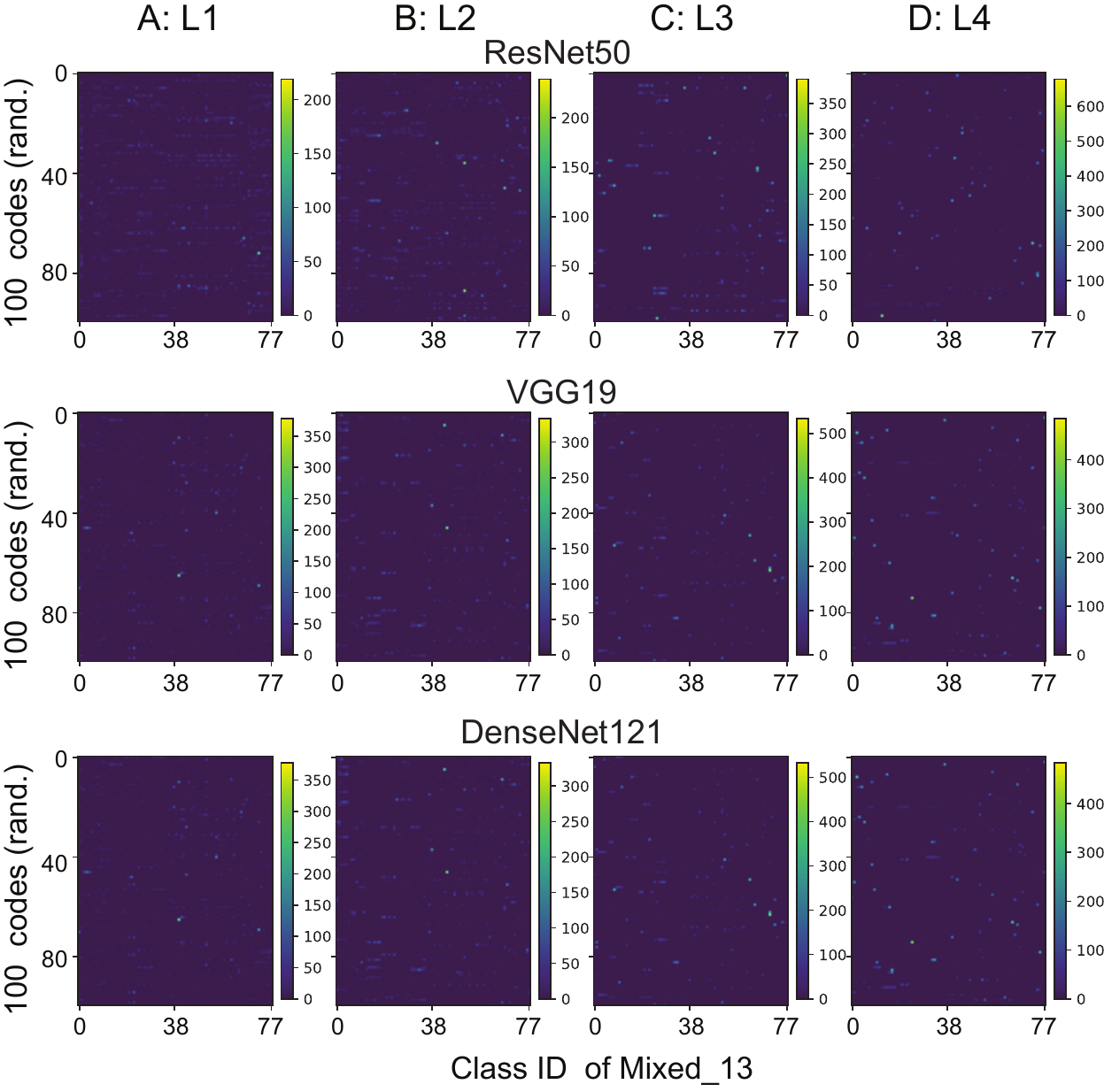}
  \caption{The correlations between symbols and classes. The top, middle and bottom rows show the correlations observed from ResNet50, VGG19 and DenseNet121. $y$-axis denotes the indices of symbols, $x$-axis denotes the class. (A)-(D), the correlations observed in layers 1-4, respectively.}
  \label{sub_fig1}
\end{figure}

\begin{figure}
  \centering
  \includegraphics[width=1\linewidth]{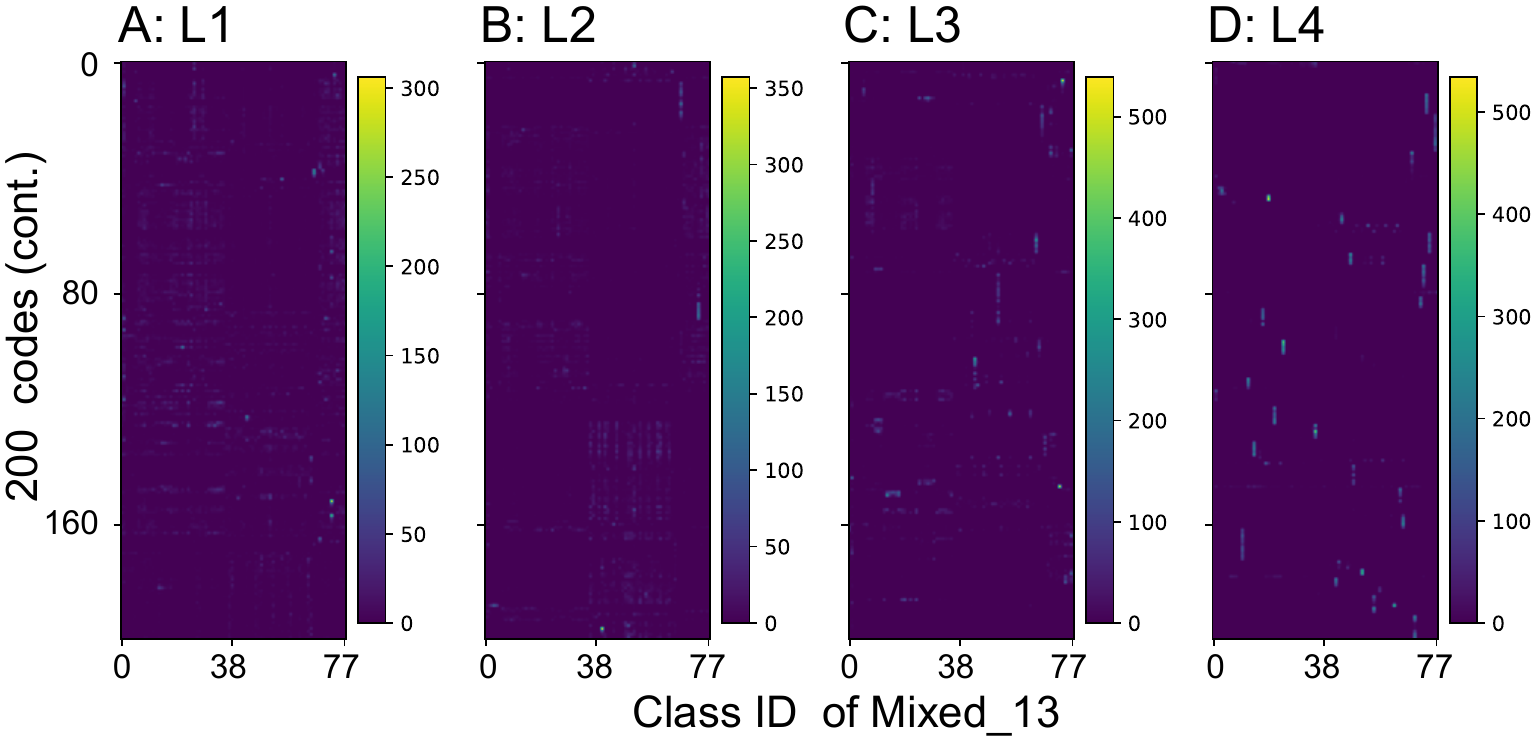}
  \caption{The correlations between symbols and classes in ResNet18. These plots show the correlation maps between 200 consecutive symbols and the labels of inputs observed in layer 1 (A), layer 2 (B), layer 3 (C) and layer 4 (D). We note that some consecutive symbols are correlated with the same classes and thus can be merged together, especially in layer 4.}
  \label{sub_fig2}
\end{figure}

If a symbol is associated with a specific class, it would appear frequently when the class object is presented, but it would not appear when other class objects are presented. Fig. \ref{correlations} shows the correlation $CM(i,j)$ between 100 randomly chosen symbols from 4 layers of ResNet18 and 78 classes. $Y$-axis denotes 100 randomly chosen symbols, and $x$-axis denotes the label (i.e., class). We found two types of symbols. First, some symbols were associated with a broad range of examples. Second, some symbols were selectively linked to specific classes. Symbols specific to classes were rare in early layers but became more common in late layers. We found equivalent results in ResNet50, VGG19 and DenseNet121 (Fig. \ref{sub_fig1}). Interestingly, we note that many symbols can be merged together (Fig. \ref{sub_fig2}),  suggesting that many symbols can be redundant. Thus, we did not increase the maximum number of clusters above 1000.  These results suggest that some symbols could be directly linked to the labels of inputs, and thus, one could infer the labels from their symbols in the hidden layers.

To test this possibility, we obtained symbols associated with test examples of Mixed\_13 and asked if they could be used to predict the labels of inputs (i.e., ROIs). In this experiment, we collected ROIs from all test images whether or not DNNs make correct predictions. From these ROIs, we collected activity vectors and obtained the test symbols by identifying the nearest cluster centers. These `test symbols' were used to predict the likelihood of each class by using $CM(i,j)$ (Eq. \ref{eq_cmap}). As we aimed to compute the probability of the class, the correlation map $CM(i,j)$ was normalized along with a row (i.e., over classes) using SoftMax (\ref{eq_probability}). By using the normalized Correlation Map $P(i,j)$, we calculated the probability $P(i,j)^m$ that $m^{th}$ ROI belonged to the class $j$ when the symbol $S_i$ was observed (Eq. \ref{eq_probability}). That is, symbol indices were used as hash keys for a look-up table.
\begin{equation}\label{eq_probability}
P(i=S_i,j)^m=\frac{e^{CM(i=S_i,j)}}{\sum_{n=1}^{78} e^{CM(i=S_i,n)}} 
\end{equation}

As we obtained 9 symbols from each ROI, we took average probability over 9 (Eq. \ref{eq_average}) and then determined the most probable class as the prediction $Pr_m$ (Eq. \ref{eq_pred}).
\begin{equation}\label{eq_average}
P(j)^m=\frac{1}{9}\Sigma_{i=1}^9 P(i,j)^m
\end{equation}

\begin{equation}\label{eq_pred}
Pr_m=argmax_j P(j)^m
\end{equation}

\begin{figure}
  \centering
  \includegraphics[width=1\linewidth]{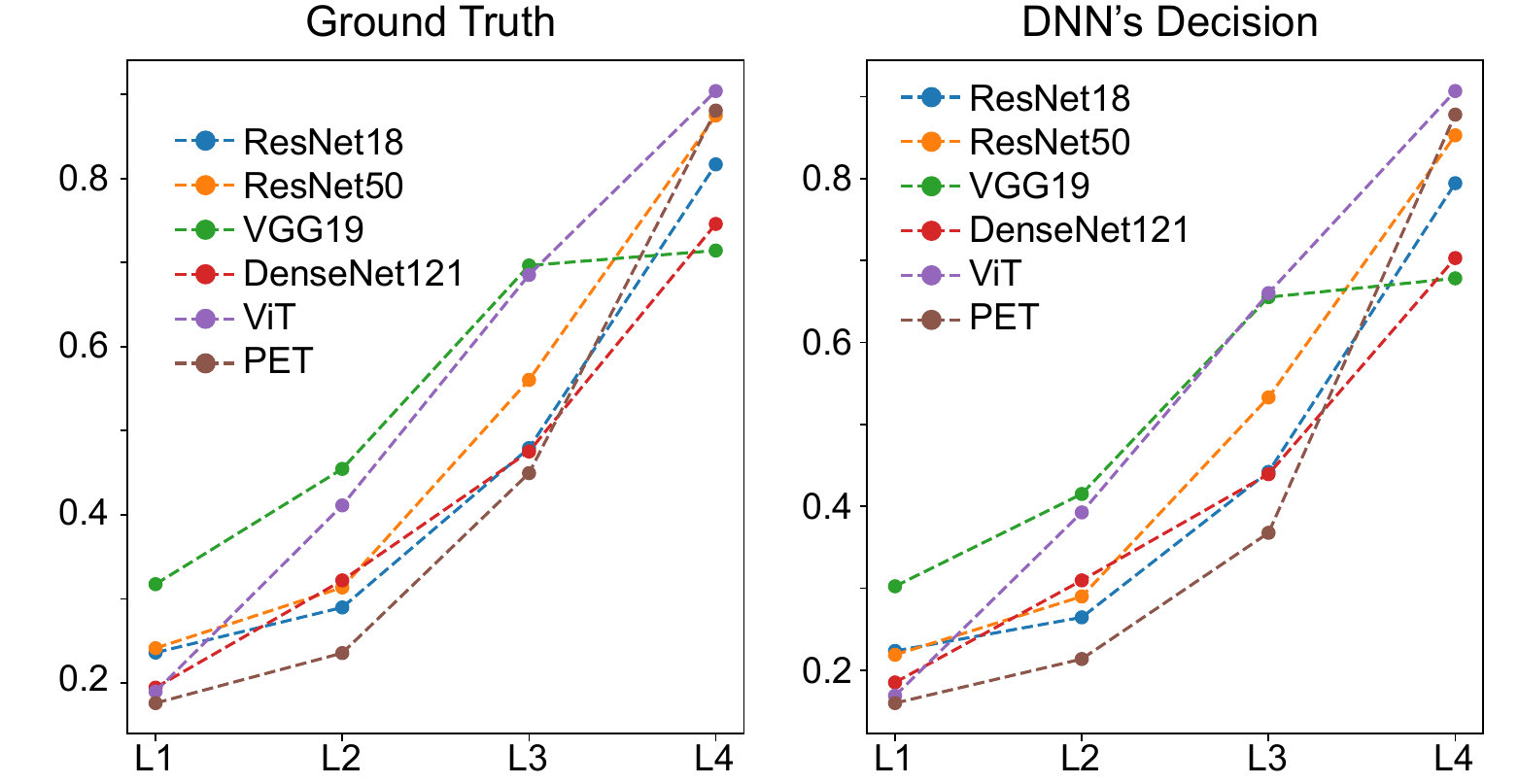}
  \caption{Symbol-based predictions. (A), Symbol-based prediction on the labels of test examples of Mixed\_13. (B), Symbol-based prediction on the DNNs' answers. PET denotes ResNet50 trained on Oxford-IIT PET. L1, L2, L3 and L4 denote the analyzed layers. For ViT, they denote $3^{rd}, 6^{th}, 9^{th}$, and $12^{th}$ embedding layer. For VGG19, they denote $2^{nd}, 3^{rd}, 4^{th}$ and $5^{th}$ Max-pooling layers. For ResNet and DenseNet, they denote all 4 layer blocks.}
  \label{correctness}
\end{figure}

Fig. \ref{correctness} shows the accuracy of the symbol-based predictions for all 5 ImageNet models and ResNet50 trained Oxford-IIT PET dataset. We note that the prediction accuracy is around 80\% in layer 4  (i.e., $5^{th}$ Max-pooling in VGG19 and $12^{th}$ embedding layer in ViT) and much lower in the earlier layers. These accuracies of ImageNet model cannot be directly compared to the models' accuracy because the choice is limited to one out of 78 classes, not 1000 classes. However, the accuracy in layer 4 and the earlier layers were significantly higher than chance (1/78), supporting that the identified symbols were associated with semantic meanings. ResNet50 trained on Oxford-IIT PET dataset also show a similar level of accuracy.

If symbols are associated with semantic meanings, they could also be associated with DNNs' predictions. We tested this possibility by correlating the symbols and DNNs' predictions on Mixed\_13 training examples (once again, 200 examples per class) instead of inputs' labels. As stated above, we used the symbols to infer their predictions on test examples and found that the symbols are predictive of DNNs' answers on test examples (Fig. \ref{correctness}B). We made equivalent observation from ResNet50 trained on Oxford-IIT PET dataset. These observations raise the possibility that DNNs could utilize symbols to make decisions.

We also tested the predictive power of symbols depending on the number of symbols. Specifically, we set a number of symbols from the 4 layers of ResNet18 to a specific number by using KMeans clustering instead of X-Means clustering. Then, we measured the predictive power of extracted symbols on labels of test examples by evaluating the accuracy of symbol-based predictions. Specifically, we tested 5 different numbers of symbols, which are 500, 1000, 1500, 2000 and 2500. As shown in Table \ref{tab_symbolnum}, the predictive power (i.e., accuracy) of symbols does not monotonically increase, as the number of symbols increases, suggesting that the predictive power of symbols is not sensitive to the exact number of symbols.

The observed link between symbols and DNNs' decisions leads us to hypothesize that symbols can be used to augment DNNs' decision-making process. Below we propose potential ways to make DNNs' operations more reliable using symbols.  
\begin{table}[]
\caption{Symbol-based predictions depending on the number of symbols. Instead of $x$-means, we set the number of symbols to be 500, 1000, 1500, 2000 and 2500. With the number of symbols fixed, we estimated the accuracy of the symbol-based predictions in 4 layers.}
\label{tab_symbolnum}
\begin{center}
\begin{tabular}{cccccc}
\hline
Layer & 500   & 1000  & 1500  & 2000  & 2500  \\ \hline
1     & 0.218 & 0.219 & 0.23  & 0.229 & 0.228 \\ \hline
2     & 0.264 & 0.27  & 0.285 & 0.28  & 0.283 \\ \hline
3     & 0.431 & 0.472 & 0.478 & 0.486 & 0.48  \\ \hline
4     & 0.782 & 0.805 & 0.812 & 0.809 & 0.819 \\ \hline
\end{tabular}
\end{center}
\end{table}

\subsection{How confident are you, DNNs?}

In the analysis above, we used $CM(i,j)$ to infer the most probable class by using the symbol indice as a hash key and $CM(i,j)$ as a look-up table. However, this look-up table can be interpreted in a completely opposite direction. If we know the label $j$ of inputs, we could infer the likelihood of individual symbols $S_i$ appearing in DNNs. With this in mind, we defined `Expected Symbol Score (ESS)' per layer with respective label/class $j$ (Eq. \ref{eq_ess}).
\begin{equation}\label{eq_ess}
ESS_j=\frac{1}{9}\Sigma_{S_1,...,S_9} P(S_i, j)^m
\end{equation}
, where $S_i$ denotes the indices of 9 symbols obtained from each ROI. That is, $ESS_j$ with respective class $j$ is the mean value of 9 probabilities from $P(i,j)^m$; see Eq. (\ref{eq_ess}).  If all observed symbols $S_i$s are exclusively correlated with class $j$,  $P(i,j)=1$, $P(i,k\neq i)=0$,  we can expect that $ESS_j=1$, and $ESS_{k\neq i}=0$. By contrast, if symbols contain no information related to classes, we can expect that $P(i,j)=ESS_j=1/78$, where 78 denotes the number of classes.  

\begin{figure}
  \centering
  \includegraphics[width=1\linewidth]{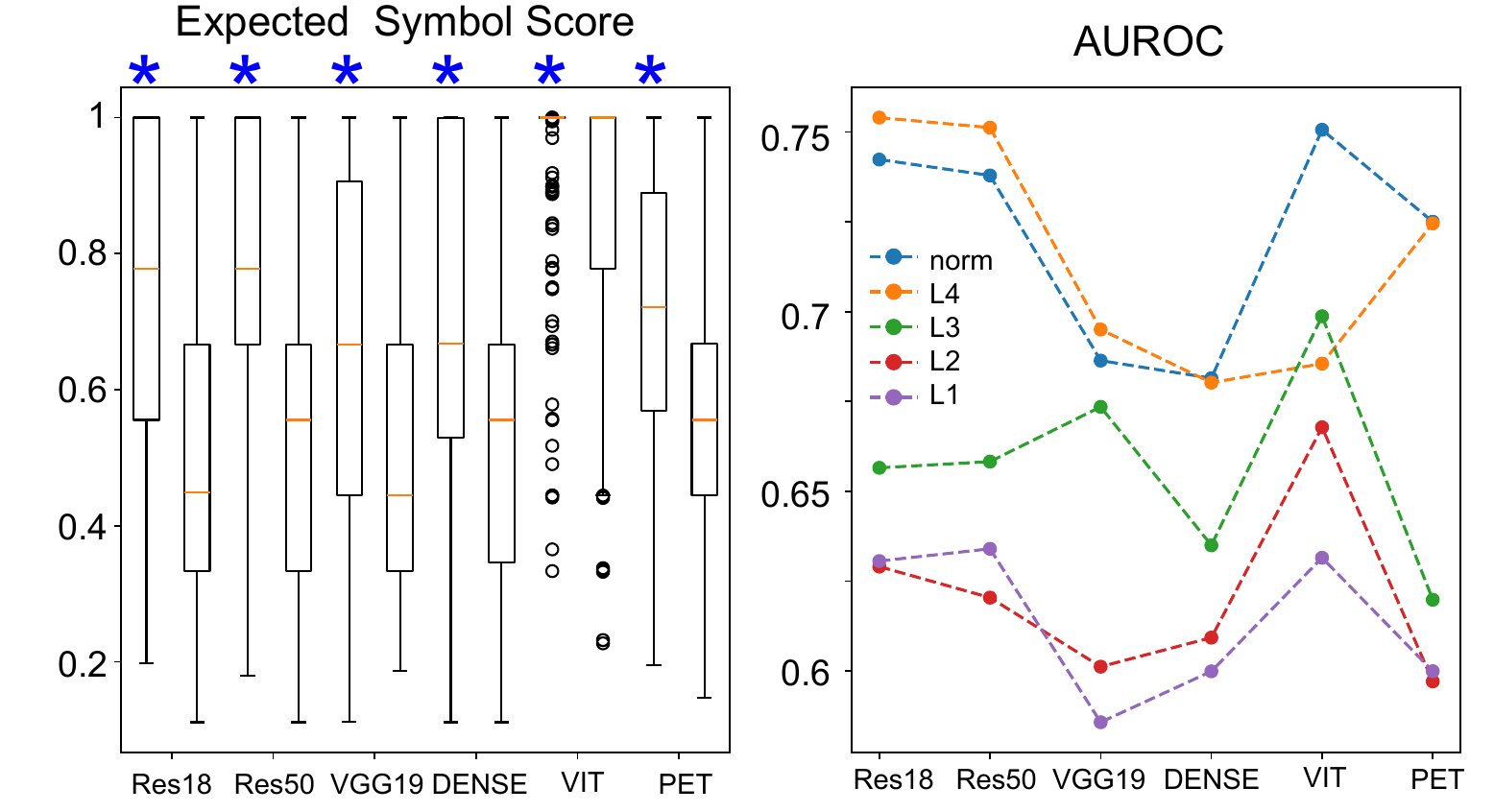}
  \caption{Symbol-based confidence level. (A), Expected symbol score (ESS) in ResNet18(Res18), ResNet50(Res50), VGG19,  DenseNet121(DENSE). ViT and ResNet50 trained on Oxford IIT-PET (PET). The asterisks denote ESSs when DNNs make correct predictions, which are higher than those when DNNs make incorrect predictions. (B), AUROC between ESS when DNNs are correct, and ESS when DNNs are wrong. We tested ESS from layer 4 (orange), layer 3 (green), layer2 (red) and layer1 (violet). Also we tested the norm from all layers, and AUROC between the norms is shown in blue.}
  \label{confidence}
\end{figure}

With this possibility in mind, we asked if ESSs obtained from layer 4 could be correlated to the accuracy of DNNs' answers. Specifically, we split ESSs into two distributions depending on DNNs’ ability to produce correct decisions. We found that ESSs were significantly higher when the predictions were correct (see data marked by * in Fig. \ref{confidence}A) than when they were incorrect (see data without * in Fig. \ref{confidence}A). To quantify the separation between correct and incorrect predictions, we estimated AUROC (the orange line in Fig. \ref{confidence}B), suggesting that ESSs indicate DNNs' likelihood of making correct predictions. We also examined ESSs from layers 1, 2 and 3 and AUROC between correct and incorrect answers (Fig. \ref{confidence}B). Further, we examined the norm of ESSs $||(ESS_1, ESS_2, ESS_3, ESS_4)||$ from all layers, and AUROC estimated from this norm were similar to those obtained from layer 4 (see blue line Fig. \ref{confidence}B). 
\subsection{Detecting out-of-distribution (OOD) examples}

Detecting OOD examples has been a central challenge in computer vision models. During training, models are forced to choose one of the predefined choices. After training, they must choose the most probable answer on any given input, even when the inputs are not related to its training. By nature, predictions on OOD examples are random and should be ignored. To address this shortcoming, OOD detection algorithms have been proposed to inform DL models (and its users) when to ignore inputs; see \cite{yang2024generalized} for a review. 

We assumed that hidden layer responses (i.e., symbols), highly optimized for in-distribution examples during training, are in synchronization across layers during inference, when inputs are drawn from in-distribution but that they are out of synchronization when inputs are drawn from out-of-distribution. Further, we speculated that ESSs could be used to measure the level of synchrony between layers and thus detect OOD examples. To test this line of hypotheses, we extracted the symbols from NINCO dataset \cite{bitterwolf2023ninco}, which contain OOD examples carefully curated against ImageNet.

In our analysis, we first used $CM(i,j)$ of layer 4 to predict the most probable class, which were incorrect, as the inputs were drawn from OOD, and tested ESSs (once again, expected symbol scores) in layers 1, 2 and 3. If the symbols are truly out of synchronization due to OOD examples, we anticipated low expected symbol scores in all three layers. We chose 18 classes from NINCO dataset, and ROIs of NINCO images were determined by GroundingDINO (an open-set segmentation model\cite{liu2023grounding}) using their real class labels. We evaluated the expected symbol scores and compared them with the ESSs of normal images from the test set of Mixed\_13. 

\begin{figure}
  \centering
  \includegraphics[width=1\linewidth]{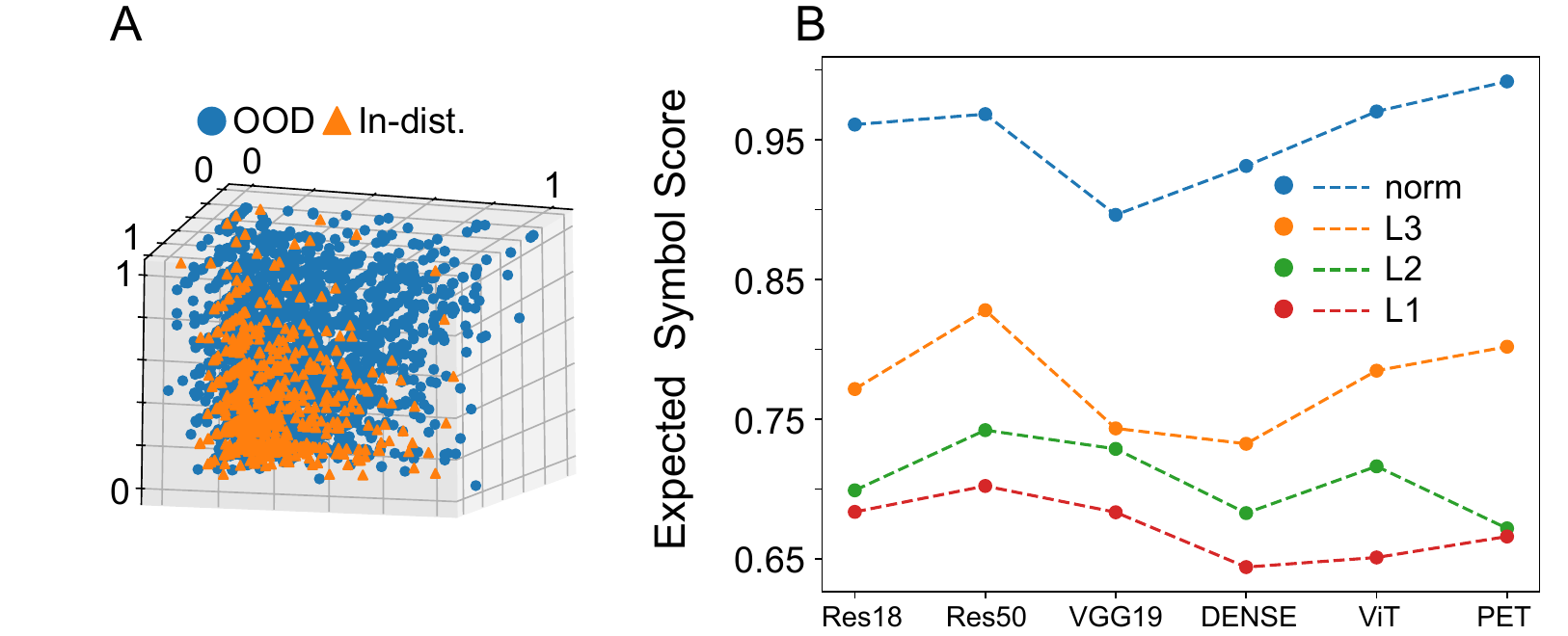}
  \caption{(A), Comparing ESSs between in-distribution and OOD examples. The orange triangles denote ESSs from layers 1,2 and 3 when inputs are drawn from OOD. The blue circles denote ESSs when inputs are drawn from in-distribution examples. (B), AUROC between ESSs for all 6 moels. AUROC were calculated using ESSs from layers 3, 2 and 1. Additionally, ESSs were also calculated using the norm of ESS from all 3 layers.}
  \label{ood_ess}
\end{figure}

Fig. \ref{ood_ess}A visualizes ESSs ($ESS_1, ESS_2, ESS_3$) from layers 1, 2, 3 of ResNet18 in a 3D space, in which $x, y, z$ denote $ESS_1, ESS_2, EES_3$, respectively. The orange circles and the blue triangles denote in-distribution and OOD examples respectively. As shown in the figure, ESSs were substantially different between in-distribution and OOD examples, and we quantified this observation by estimating AUROC for all 6 models. The estimated AUROC from ESSs of all 3 layers and their norms $||(ESS_1, ESS_2, ESS_3)||$ suggest that ESSs are effectively separated between normal and OOD examples and can be used to detect OOD examples.

\subsection{Symbols used to address the vulnerabilities to adversarial perturbations}
Since the discovery of DL models' vulnerability to adversarial examples \cite{reviewadver1, reviewadver2}, adversarial perturbation has emerged as one of the greatest threats to DL safety. Given that all inputs are progressively processed in modern DNNs, it seems natural to assume that adversarial perturbation may change internal representation on layer-by-layer basis.  An earlier study \cite{umap_adver} found that the difference in representations in the early layers was not noticeable, but in the late layers it grew significantly. We asked if ESSs can be used to detect the influence on adversarial perturbation. Specifically, we crafted adversarial examples from validation images of Mixed\_13 using AutoAttack \cite{autoattack}, which is known to be one of the most effective attacks. In the experiment, we used a `standard' mode with $eps=0.03$ and $norm=Linf$. For individual inputs, we used labels of clean images to detect ROIs and estimated the corresponding symbols; that is, ROIs are identified with original labels, not the manipulated predictions. 

\begin{figure}
  \centering
  \includegraphics[width=1\linewidth]{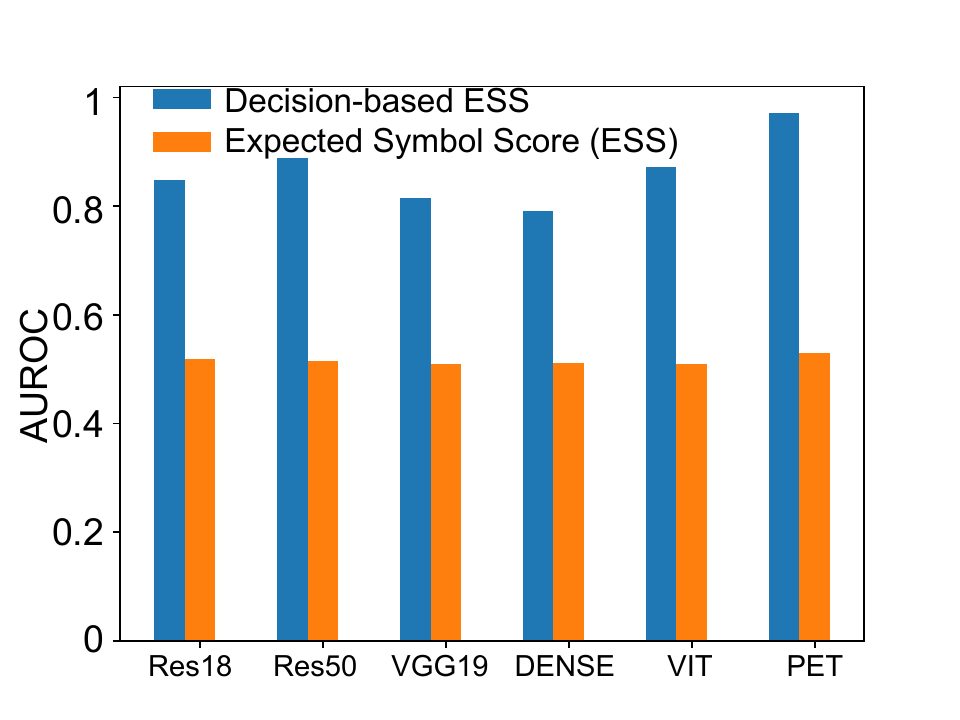}
  \caption{Symbol-based detection of adversarial perturbations. We estimated ESS using models' decisions and the predictions of final layer (i.e., L4). Then, we tested if ESS are significantly different between normal and adversarial inputs. The orange line denotes AUROC calculated from decision-based ESS, whereas the blue dashed line denotes AUROC calculated ESS. The green line denotes the accuracy of symbol-based predictions on the adversarial inputs.}
  \label{adversarial}
\end{figure}

For ROIs, we estimated the norm of ESSs from all 4 layers $||(ESS_1, ESS_2, ESS_3, ESS_4)||$ and calculated AUROC by comparing the norms of ESSs between normal and adversarial inputs. We note that the AUROC estimated from the norms were close to 0.5 (see the orange bars in Fig. \ref{adversarial}), suggesting that the symbols in DNNs were not significantly different between normal and adversarial examples. Thus, we tested an alternative method. So far, ESSs have been computed using the most probable class predicted by the layer 4, which is the closest to the logit/output layer in the model. Specifically, we used layer 4' s  symbols ($S_i$) and correlation map $CM(i,j)$ to predict a class label $j$, but ESSs can also be computed using DNNs' predictions directly. That is, we can set $j$ to be the model's prediction. 

If ESSs are computed with DNNs' predictions, we can effectively evaluate the consistency between models' decisions and internal symbols, which we assume could be useful in detecting adversarial perturbation. Notably, we analyzed unmodified ImageNet models, whose output nodes correspond to 1,000 classes, not 78 of Mixed\_13. As we did not set the target class during adversarial attacks, the predictions were not necessarily confined to 78 classes of Mixed\_13. Thus, to build ESSs using their predictions, we selected adversarial and normal symbols when predictions are one of 78 classes in Mixed\_13 (see Table \ref{tab_roi} for the number of ROIs included in this analysis).  For ResNet50 trained on Oxford-IIT PET dataset (shown as `PET' in the $x$-axis), we used all perturbed inputs.  Decision-based ESSs are compared between normal and adversarial inputs (see the blue bars in Fig. \ref{adversarial}). As shown in the figure, `decision-based ESSs' are significantly different between normal and adversarial inputs in all 6 models, suggesting that ESSs with respect to DNNs' (manipulated) predictions can be used to detect adversarial perturbation. 

\begin{figure}
  \centering
  \includegraphics[width=1\linewidth]{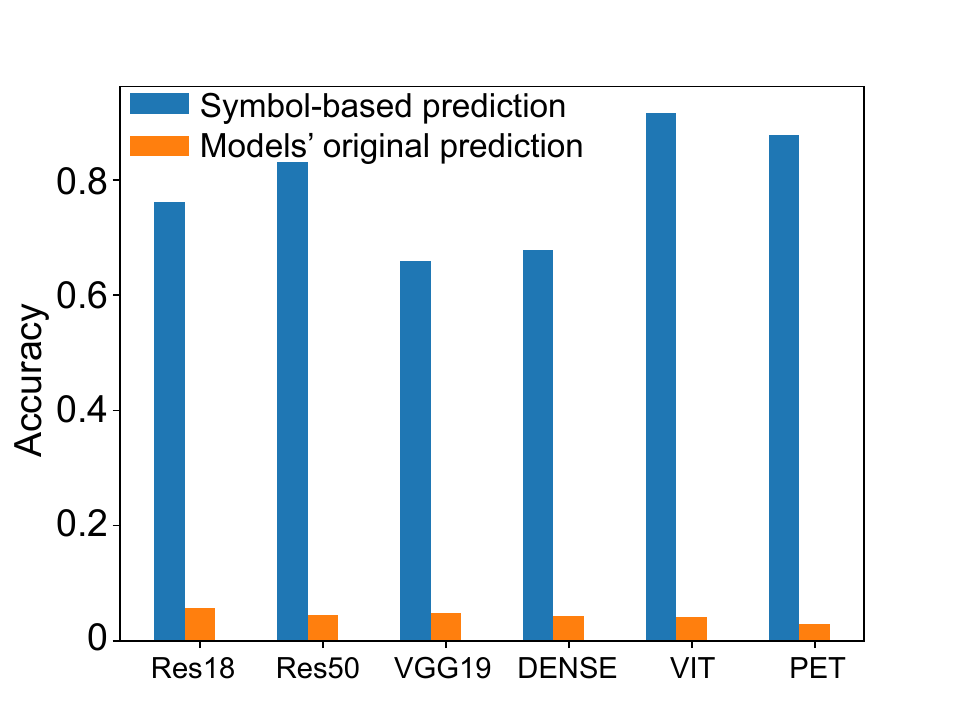}
  \caption{The accuracy of symbol-based predictions on adversarial inputs perturbations. We first made predictions using the symbols in final layer (i.e., L4) and evaluated their accuracy. Blue bars show that the symbols in L4 can accurately predict the original classes of adversarial inputs. For comparison, we also display the accuracy of DL models' original predictions (see orange bars).}
  \label{adversarial_acc}
\end{figure}

Further, inspired by the results suggesting that symbols can be used to predict labels of inputs  (Fig. \ref{correctness}), we made predictions on adversarial inputs using symbols in L4 (e.g., $12^{th}$ embedding layer for ViT). Fig. \ref{adversarial_acc}, compares the accuracy of symbol-based predictions and the accuracy of the models' original predictions on adversarial inputs. We note that even though adversarial inputs effectively lowered the accuracy of DL models' predictions (see orange bars), their impact on the symbol-based predictions was minimal. This result indicates that symbols can be used to address DL models' vulnerabilities to adversarial perturbations. 

\begin{table}[]
\caption{Number of normal and adversarial ROIs. For both normal and adversarial inputs, we first selected inputs (i.e., images), on which DNNs' predictions were one of 78 classes of Mixed\_13. Then, we selected all ROIs from these examples (both normal and adversarial). We list the number of ROIs selected below. It should be noted that images are not chosen using the correctness of DNNs' predictions.}
\label{tab_roi}

\begin{center}
\begin{tabular}{ccc}

\hline
Model       & Normal & Adversarial \\ \hline
ResNet18    & 4002   & 1161        \\ \hline
ResNet50    & 4320   & 1432        \\ \hline
VGG19       & 3989   & 1015        \\ \hline
DenseNet121 & 4133   & 765         \\ \hline
VIT         & 4419   & 549         \\ \hline
\end{tabular}
\end{center}
\end{table}

\subsection{Temporary learning}
\begin{figure}
  \centering
  \includegraphics[width=1\linewidth]{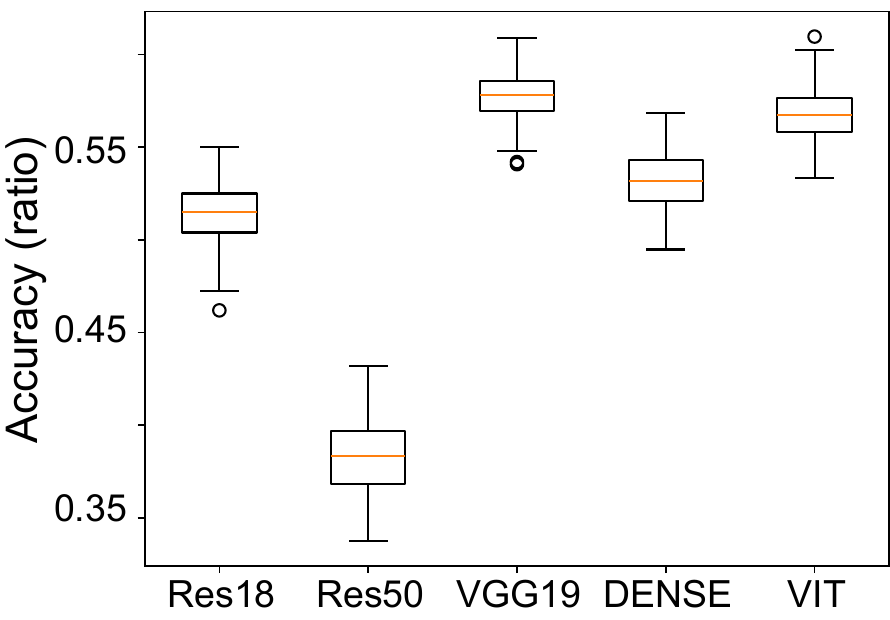}
  \caption{The accuracy of predictions on the OOD examples for ResNet18(Res18), ResNet50(Res50), VGG19, DenseNet121(DENSE) and ViT. We resampled the training and test examples from 18 classes of NINCO dataset \cite{bitterwolf2023ninco} for 100 times and collected their accuracies.
  }
  \label{temporay_ood}
\end{figure}

One of DL's shortcomings is the lack of continual learning ability, whereas the brain can learn continuously. The brain can use previously obtained knowledge and incorporate new knowledge, but DL models' catastrophic forgetting, the deletion of previously obtained information to learn the new, prevents continual learning. Since symbols can serve as previous knowledge, we speculate that the temporary mapping between symbols of in-distribution examples and OOD examples may allow a short-term (temporary) learning of OOD examples, which can serve as a stop-gap measure, until a more formal retraining becomes available. To address this possibility, we randomly split NINCO symbols (obtained from 18 classes) into 2 distinct distributions, training sets and test sets. The size of the training and test sets are half the size of NINCO symbols. Then, the training symbols were used to construct $CM(i,j)$, where class label $j$ is one of 18 classes of NINCO dataset. Then, we asked if the test symbols can predict the inputs' classes. As training/test symbols are selected randomly, we resampled them 100 times and reported the results from all 100 experiments (Fig. \ref{temporay_ood}). We observed notable differences between the models. More importantly, the accuracy is around 40-50\% (Fig. \ref{temporay_ood}) for all 6 models, which is far higher than the random guess ($1/18$), supporting that internal symbols can be used for short-term learning. 

\subsection{Symbols associated with internal features}
In this study, we extract symbols from ROIs. This is based on the assumption that classifiers rely on internal features of visual objects. We tested this assumption by creating cropped versions of ROIs (containing objects) and evaluated DL models' accuracy on full images, ROIs and cropped ROIs. Specifically, the three cropped versions, ROI81, ROI64 and ROI49\%, contain 81\%, 64\% and 49\% of pixels of ROIs obtained from STCert. All cropped ROIs and the full ROI are aligned to have the same centers, and we tested DL models’ accuracy on the test set of Mixed\_13. The accuracy on ROI is normalized to the accuracy on full images. We note that the drop in the accuracy on ROIs is limited even at ROI:49\% (Table \ref{tab_acc}), suggesting that DL models can indeed learn to use internal features of visual objects, which supports our assumption.
 
\begin{table}[]
\caption{We tested the accuracy of 6 ImageNet models, ResNet18(RES18), ResNet50(RES50), VGG19, DenseNet121(DENS121) and VIT. }
\label{tab_acc}

\begin{center}
\begin{tabular}{llllll}
\hline
Model  & Full & ROI   & ROI81 & ROI64 & ROI49 \\ \hline
RES18  & 1    & 0.851 & 0.835 & 0.819 & 0.79  \\ \hline
RES50  & 1    & 0.896 & 0.88  & 0.87  & 0.852 \\ \hline
VGG19  & 1    & 0.839 & 0.817 & 0.799 & 0.773 \\ \hline
DEN121 & 1    & 0.87  & 0.848 & 0.829 & 0.813 \\ \hline
VIT    & 1    & 0.896 & 0.879 & 0.868 & 0.851 \\ \hline
\end{tabular}
\end{center}
\end{table}

\section{Discussion}

We probed the existence of internal symbols closely linked to DNNs' decision making. More specifically, we combined STCert and traditional machine learning algorithms to detect internal symbols. Our analysis of these internal symbols suggest that they can be used to 1) predict levels of accuracies of DNNs' answers  (Fig.\ref{correctness}), 2) detect OODs (Fig.\ref{ood_ess}) and adversarial inputs  (Fig.\ref{adversarial}), 3) make robust predictions on adversarial inputs (Fig. \ref{adversarial_acc}) and 4) enable  temporary learning of OOD examples (Fig.\ref{temporay_ood}). Based on our analysis, we propose that internal symbols could play a crucial role in addressing the shortcomings of  DNNs' and helping us build more reliable and safer DL models. 

\subsection{Limitations}

We should underline that the purpose of our study is not to provide a comprehensive survey of methods that can extract internal symbols and potential utilities of internal symbols. Instead, we focused on searching for positive evidence of the existence of internal symbols associated with DNNs' decisions and exploring their applications.

\bibliography{main}

\bibliographystyle{unsrt}

\clearpage
\setcounter{page}{1}

\setcounter{table}{0}
\renewcommand{\tablename}{Supplementary Table}

\setcounter{figure}{0}
\renewcommand{\figurename}{Supplementary Figure}

\section{Appendix and Supplementary Material}
\subsection{STCert implementation}\label{stcert_det}
The two versions of STCert were discussed in the original study \cite{lee2023having}. The naive STCert uses the bounding boxes returned by the segmentation models as ROIs. The context-aware STCert adds some background to the bounding boxes to create ROIs. As the context-aware STCert is more robust, we adopted the context-aware STCert in this study. The workflow of STCert can be summarized as follows:

\begin{enumerate}
    \item Detect bounding boxes.
    \item Create 5 candidates of bounding boxes by enlarging ROIs. The first candidate is ROI itself.
    \item Crop pixels within the candidates and forward them into DNNs. That is, we have the list of 5 second thought predictions.
    \item Examine if the original prediction is included in the list of second predictions. 
    \item Return the bounding box as ROI, if the original prediction is in the list. If not, ignore the bounding box and return Null. 
\end{enumerate}

\subsection{Potential links to the brain structure}

Our analysis raises the possibility that internal symbols can be used to augment DNNs' decision-making process. It may even be possible for a second network to use symbols to correct responses in each layer locally during inference. When urgent and immediate decisions are demanded, DNNs' decisions can be used, as they are, but for more reliable answers, a second network may use symbols to modify hidden layers' responses. This may remind some readers of thalamo-cortical loops in the brain. Our analysis has been inspired by thalamo-cortical loops and indicates that thalamo-cortical loop-like error correction could be constructed using internal symbols to build more reliable DNNs.

\begin{table}[]
\caption{List of classes in the Mixed\_13.}
\label{tab_class}
\begin{center}
\begin{tabular}{lc}
\hline
\textbf{Category} & \textbf{Class}                                                                                                                                \\ \hline
Fish              & \begin{tabular}[c]{@{}c@{}}`tench', `goldfish', `great\_white\_shark', \\ `tiger\_shark', `hammerhead', `electric\_ray'\end{tabular}          \\ \hline
Bird              & \begin{tabular}[c]{@{}c@{}}`cock', `hen', `ostrich', \\ `brambling', `goldfinch', `house\_finch'\end{tabular}                                 \\ \hline
Dog               & \begin{tabular}[c]{@{}c@{}}`Chihuahua', `Japanese\_spaniel', `Maltese\_dog',\\  `Pekinese', `Shih-Tzu', `Blenheim\_spaniel'\end{tabular}      \\ \hline
Carnivore         & \begin{tabular}[c]{@{}c@{}}`cougar', `lynx', `leopard', \\ `snow\_leopard', `jaguar', `lion'\end{tabular}                                     \\ \hline
Insect            & \begin{tabular}[c]{@{}c@{}}`tiger\_beetle', `ladybug',`ground\_beetle',\\  `long-horned\_beetle', `leaf\_beetle', 'dung\_beetle'\end{tabular} \\ \hline
Primate           & \begin{tabular}[c]{@{}c@{}}`guenon', `patas', `baboon',\\  `macaque', `langur', `colobus'\end{tabular}                                        \\ \hline
Car               & \begin{tabular}[c]{@{}c@{}}`ambulance', `cab', `convertible',\\  `jeep', `limousine', `station\_wagon'\end{tabular}                           \\ \hline
Furniture         & \begin{tabular}[c]{@{}c@{}}`barber chair', `bassinet', `bookcase', \\`chiffonier', `china\_cabinet', `cradle'\end{tabular}                   \\ \hline
Computer          & \begin{tabular}[c]{@{}c@{}}`desktop\_computer', `hand-held\_computer',\\  `laptop', `notebook', `slide\_rule', `web\_site'\end{tabular}       \\ \hline
Fruit             & \begin{tabular}[c]{@{}c@{}}`Granny\_Smith', `rapeseed',  `corn', \\ `acorn', `hip', `buckeye'\end{tabular}                                     \\ \hline
Fungus            & \begin{tabular}[c]{@{}c@{}}`coral\_fungus', `agaric',`gyromitra',\\  `stinkhorn', `earthstar', `hen-of-the-woods'\end{tabular}                \\ \hline
Truck             & \begin{tabular}[c]{@{}c@{}}`fire\_engine', `garbage\_truck', `moving\_van',\\  `pickup', `police\_van', `tow\_truck'\end{tabular}             \\ \hline
\end{tabular}
\end{center}
\end{table}

\end{document}